# An Uncertainty Framework for Classification

**Loo-Nin Teow**    **Kia-Fock Loe**
School of Computing
National University of Singapore
Science Drive 2, S 117559, Singapore

## Abstract

We define a generalized likelihood function based on uncertainty measures and show that maximizing such a likelihood function for different measures induces different types of classifiers. In the probabilistic framework, we obtain classifiers that optimize the cross-entropy function. In the *possibilistic* framework, we obtain classifiers that maximize the interclass margin. Furthermore, we show that the support vector machine is a sub-class of these maximum-margin classifiers.

## 1 INTRODUCTION

Uncertainty is a natural and unavoidable part of pattern classification in real-world domains. Given a sample that has never before been encountered, a certain amount of uncertainty is always involved in deciding the class to be assigned to that sample. We make the decision associated with the least uncertainty, based on past encounters with other samples. It is therefore necessary to represent and deal with such uncertainty in an effective manner.

In much of the literature on classification, this uncertainty is typically represented as probabilities (Devroye et al. 1996; Jurgen 1996; Langley et al 1992), since probability theory is already well established. In the Bayesian context, we choose the class with the maximum posterior probability. Hence, traditional pattern recognition often involves finding suitable functions to model probability distributions, and manipulating the outputs of such functions on the premise that they are true probabilities.

However, as we shall argue in this paper, such an assumption may not be valid, and it may be more appropriate to treat the uncertainty in a *possibilistic* manner (Bezdek et al 1999; Klir & Yuan 1995; Tanaka & Guo 1999). To this end, we propose an uncertainty framework for classification, of which the probabilistic and possibilistic frameworks are special cases. As we shall show later, the probabilistic framework leads to classifiers that optimize the cross-entropy function. On the other hand, the possibilistic framework leads to maximum-margin classifiers, which have been shown to exhibit superior generalization in the recent literature (Scholkopf 1997; Yang & Liu 1999).

We begin by discussing the probabilistic framework in the next section. We will then show how this framework can be generalized to handle other measures of uncertainty. In particular, we will show how to induce possibilistic classifiers. Next, we will focus on the two-class problem for the linear classifier, and show its relation to the support vector machine (Burges 1998; Cortes & Vapnik 1995; Vapnik 1995). Finally, we will discuss the extension to non-linear classifiers.

## 2 THE PROBABILISTIC FRAMEWORK

Given $m$ classes, let $\mathbf{x} \in \mathbf{R}^d$ be a $d$-dimensional input vector and $y \in \{1, ..., m\}$ be the class label. Consider a set of independent and identically distributed (i.i.d.) training samples $\{\mathbf{x}_i, y_i\}$ for $i = 1$ to $n$. In pattern classification, we are interested in the class posterior (conditional) probabilities $p(y \mid \mathbf{x}, \theta)$, where $\theta$ is the set of parameters for the corresponding probability distribution. The conditional probability that an input vector $\mathbf{x}_i$ belongs to class h, for example, is $p(h \mid \mathbf{x}_i, \theta_h)$.

One popular way of estimating the conditional probability distributions is to first estimate the joint probabilities, i.e., $p(\mathbf{x}, y \mid \theta)$, and then condition them using Bayes rule. The estimation is typically done by maximizing the joint likelihood for each class using the EM algorithm (Dempster et al. 1977). An alternative way is to treat the class as a random variable in a single distribution (e.g. a Gaussian or multinomial distribution) and then maximize the likelihood for this distribution, as done for example in (Jordan & Jacobs 1995). In this paper, we formulate the conditional likelihood in such a way that allows it to be generalized to other uncertainty measures.

The *total conditional likelihood* for class $h$ is the total conditional probability of observing $y_i = h$, i.e.,

$$L_h = \prod_{i \mid y_i = h} p(h \mid \mathbf{x}_i, \theta_h) \qquad [1]$$

The product operation is due to the i.i.d. assumption. We can interpret this likelihood as a measure of the certainty that all the samples with $y_i = h$ belong to class $h$. To maximize the certainty, we need to find the parameters such that this likelihood is maximized. Hence, we want to maximize the total conditional likelihood, or equivalently its logarithm, for *each* class with respect to the parameters, i.e., maximize the likelihood function

$$L = \sum_{h}^{m} \log L_h \qquad [2]$$

Substituting [1] into [2], we get the probabilistic likelihood function:

$$L = \sum_{h}^{m} \log \prod_{i|y_i=h} p(h|\mathbf{x}_i, \theta_h)$$

$$= \sum_{h}^{m} \sum_{i|y_i=h} \log p(h|\mathbf{x}_i, \theta_h) \qquad [3]$$

$$= \sum_{i}^{n} \sum_{h}^{m} q_{hi} \log p(h|\mathbf{x}_i, \theta_h)$$

where

$$q_{hi} = \begin{cases} 1 & \text{if } y_i = h \\ 0 & \text{otherwise} \end{cases} \qquad [4]$$

This is exactly the negative of the cross-entropy function often employed in the training of neural network classifiers (Baum & Wilczek 1988; Hertz et al. 1991; Hinton 1989), although it is derived here under a different context. In the cross-entropy function, $q_{hi}$ are the target posterior probabilities, and the objective is to minimize the Kullback-Leibler divergence of the posterior probability distribution from the target distribution. In the above formulation, $q_{hi}$ are meant to index the probabilities with respect to the class, but can also be interpreted as posterior probabilities as in the cross-entropy function. In this paper, we shall call any classifier that minimizes the cross-entropy function a probabilistic classifier.

As an example, we can model the posterior probabilities using $m$ linear classifiers. Each linear classifier implements a hyperplane that separates its corresponding class from the other classes. The equation of each hyperplane is

$$f_h(\mathbf{x}) = \mathbf{w}_h \cdot \mathbf{x} + b_h \qquad [5]$$

where $\mathbf{w}_h$ is a weight vector and $b_h$ is a bias term. Each posterior probability is typically computed using the *softmax* function:

$$p(h|\mathbf{x}_i, \theta_h) = \frac{\exp(f_h(\mathbf{x}_i))}{\sum_{k}^{m} \exp(f_k(\mathbf{x}_i))} \qquad [6]$$

Hence, the parameter set $\theta_h$ is $\{\mathbf{w}_h, b_h\}$. Substituting the posterior probabilities into the cross-entropy function and using gradient descent to optimize it with respect to the parameters leads to the well-known perceptron learning rule (Rosenblatt 1962).

Alternatively, we can model the posterior probabilities by the outputs of a multi-layered perceptron network. In this case, maximizing $L$ by gradient descent leads to the backpropagation algorithm (Rumelhart et al 1986).

## 3 THE POSSIBILISTIC FRAMEWORK

In this section, we explore the extension of the likelihood function to other forms of uncertainty. In general, the measure $p(h|\mathbf{x}_i, \theta_h)$ represents the amount of uncertainty in associating class $h$ with $\mathbf{x}_i$, and probability is only one way to represent this uncertainty. Instead of probability, it can be the belief or confidence that the class of $\mathbf{x}_i$ is $h$, or it can be the degree to which $\mathbf{x}_i$ belongs to class $h$. All of these can be expressed in terms of *possibilities* (Bezdek et al 1999; Klir & Yuan 1995; Tanaka & Guo 1999).

Both probability and possibility values range from 0 to 1. However, one key difference is that while probability distribution values sum up to 1, this is not a requirement for possibilistic distributions. In the case of conditional probabilities, the following condition must hold:

$$\sum_{h}^{m} p(h|\mathbf{x}_i, \theta_h) = 1 \qquad [7]$$

In the possibilistic case, the above condition is not mandatory.

For uncertainty measures in general, the product operation as used in [1] may not be appropriate. We need to generalize it to conjunction operators that are better suited for handling a wide range of uncertainty measures. Hence, the total conditional likelihood for class $h$ can be generalized as follows

$$L_h = \bigotimes_{i|y_i=h} p(h|\mathbf{x}_i, \theta_h) \qquad [8]$$

where $\otimes$ is a conjunction operator which can be any t-norm (Gupta & Qi 1991). The generalized likelihood function to be maximized is then

$$L = \sum_{h}^{m} \log \bigotimes_{i|y_i=h} p(h|\mathbf{x}_i, \theta_h) \qquad [9]$$

In the case of possibilistic values, the standard conjunction operator is *min*. The *min* operator is especially suitable because it is non-interactive and hence assumes the least about the nature of the uncertainties. The possibilistic likelihood function is then

$$L = \sum_{h}^{m} \log \min_{i|y_i=h} p(h|\mathbf{x}_i, \theta_h) \qquad [10]$$

In this paper, we shall call any classifier that maximizes the above criterion a possibilistic classifier.

An interesting special case arises in two-class linear discrimination, which we shall now study in detail in the next section.

### 3.1 TWO-CLASS LINEAR DISCRIMINATION

For two classes, we redefine the class labels as +1 and -1, i.e. $y \in \{+1, -1\}$. Only one hyperplane $f(\mathbf{x})$ is required, and we define

$$p(+1|\mathbf{x}_i, \theta_h) = g(f(\mathbf{x}_i))$$
$$p(-1|\mathbf{x}_i, \theta_h) = g(-f(\mathbf{x}_i))$$
[11]

where, given the weight vector $\mathbf{w}$ and the bias term $b$,

$$f(\mathbf{x}) = \mathbf{w} \cdot \mathbf{x} + b \quad [12]$$

and $g$ is the sigmoid or logistic function

$$g(z) = \frac{1}{1+e^{-z}} \quad [13]$$

We impose the constraint that the weight vector is normalized to unit length, i.e., $|\mathbf{w}| = 1$. This gives $f(\mathbf{x})$ a geometric interpretation: its magnitude is the Euclidean distance of $\mathbf{x}$ from the hyperplane. The constraint can be easily incorporated via a Lagrange multiplier $\lambda$. Thus the possibilistic likelihood function becomes

$$L = \log \min_{i|y_i=+1} g(f(\mathbf{x}_i)) + \log \min_{i|y_i=-1} g(-f(\mathbf{x}_i)) - \lambda(\mathbf{w} \cdot \mathbf{w} - 1) \quad [14]$$

Bearing in mind that the sigmoid function is monotonic increasing and that $g(-z) = 1 - g(z)$, this criterion can be equivalently written as

$$L = \log \min_{i|y_i=+1} g(f(\mathbf{x}_i)) +$$
$$\log\left(1 - \max_{i|y_i=-1} g(f(\mathbf{x}_i))\right) - \lambda(\mathbf{w} \cdot \mathbf{w} - 1)$$
$$= \log g\left(\min_{i|y_i=+1} f(\mathbf{x}_i)\right) +$$
$$\log\left(1 - g\left(\max_{i|y_i=-1} f(\mathbf{x}_i)\right)\right) - \lambda(\mathbf{w} \cdot \mathbf{w} - 1) \quad [15]$$
$$= \log g\left(\min_{i|y_i=+1} (\mathbf{w} \cdot \mathbf{x}_i) + b\right) +$$
$$\log\left(1 - g\left(\max_{i|y_i=-1} (\mathbf{w} \cdot \mathbf{x}_i) + b\right)\right) - \lambda(\mathbf{w} \cdot \mathbf{w} - 1)$$

The maximization of $L$ can be better understood from the following theorems.

**Theorem 1.**

When $L$ (as defined above) is maximized, the following equalities hold:

(a) $b = -\frac{1}{2}\left(\min_{i|y_i=+1}(\mathbf{w} \cdot \mathbf{x}_i) + \max_{i|y_i=-1}(\mathbf{w} \cdot \mathbf{x}_i)\right)$

(b) $\min_{i|y_i=+1} f(\mathbf{x}_i) = -\max_{i|y_i=-1} f(\mathbf{x}_i)$

(c) $\min_{i|y_i=+1} g(f(\mathbf{x}_i)) = \min_{i|y_i=-1} g(-f(\mathbf{x}_i))$

□

Part (a) of Theorem 1 can be proven easily by finding the derivative of $L$ with respect to $b$, equating this derivative with 0, and solving for $b$. Parts (b) and (c) follow logically from part (a). The theorem is stating that the optimal hyperplane always lies exactly halfway between the boundaries of both classes. This forms the basis of the following theorems.

**Theorem 2.**

If the training data is linearly separable, then when $L$ is maximized, the following propositions are true:

(a) $\min_{i|y_i=+1} f(\mathbf{x}_i) > 0$ and $\max_{i|y_i=-1} f(\mathbf{x}_i) < 0$

(b) The interclass margin
$$\left(\min_{i|y_i=+1} f(\mathbf{x}_i) - \max_{i|y_i=-1} f(\mathbf{x}_i)\right)$$ is maximized.

□

Part (a) of Theorem 2, which follows from the definition of linear separability, is simply stating that when the training data is separable, all positive and negative samples must lie on opposite sides of the hyperplane. Part (b), an important result, follows from the application of Theorem 1(b, c) and the fact that both the logarithm and sigmoid functions are monotonic increasing.

**Theorem 3.**

If the training data is linearly non-separable, then when $L$ is maximized, the following propositions are true:

(a) $\min_{i|y_i=+1} f(\mathbf{x}_i) \leq 0$ and $\max_{i|y_i=-1} f(\mathbf{x}_i) \geq 0$

(b) The interclass overlap
$$\left(\max_{i|y_i=-1} f(\mathbf{x}_i) - \min_{i|y_i=+1} f(\mathbf{x}_i)\right)$$ is minimized.

□

Theorem 3 can be proven in a similar way to Theorem 2.

In short, the possibilistic linear classifier seeks a hyperplane that maximizes its displacement from the boundaries of both classes. For linearly separable data, maximizing $L$ serves to *maximize the interclass margin*. In the non-separable case, maximizing $L$ serves to *minimize the interclass overlap*.

We can also describe the maximization of $L$ conceptually as follows. The first term of $L$ in [13] corresponds to the positive samples furthest in the negative direction (with respect to the weight vector), while the second term corresponds to the negative samples furthest in the positive direction. Since the magnitude of $f(\mathbf{x})$ is the Euclidean distance of $\mathbf{x}$ from the hyperplane, maximizing $L$ "pushes" both sets of boundary samples apart in opposite directions along the weight vector.

Finding the hyperplane that maximizes the interclass margin is also the same strategy adopted by a class of pattern classifiers known as support vector machines. In the next section, we show the equivalence between the possibilistic linear classifier and the support vector machine.

## 4 SUPPORT VECTOR MACHINES

In the past few years, there has been increasingly widespread interest in a class of pattern classifiers known as support vector machines (SVMs), due to their strong theoretical foundations, the existence of global solutions, and most importantly, good generalization performance (Burges 1998; Cortes & Vapnik 1995; Scholkopf 1997; Vapnik 1995; Yang & Liu 1999).

In a linear SVM trained on linearly separable data, $f(\mathbf{x})$ as defined in [12] is normalized such that

$$f(\mathbf{x}_i) \geq +1 \quad \text{for } y_i = +1$$
$$f(\mathbf{x}_i) \leq -1 \quad \text{for } y_i = -1$$
[16]

The margin width is then equal to

$$\min_{i|y_i=+1} \frac{f(\mathbf{x}_i)}{|\mathbf{w}|} - \max_{i|y_i=-1} \frac{f(\mathbf{x}_i)}{|\mathbf{w}|}$$
$$= \frac{1}{|\mathbf{w}|} \left( \min_{i|y_i=+1} f(\mathbf{x}_i) - \max_{i|y_i=-1} f(\mathbf{x}_i) \right)$$
$$= \frac{2}{|\mathbf{w}|}$$
[17]

That is, the margin width in this case is inversely proportional to the weight magnitude.

The SVM maximizes the margin width indirectly by minimizing $|\mathbf{w}|$ subject to the constraints in [16]. This can be reformulated as the following quadratic programming criterion to be maximized:

$$\sum_i^n \alpha_i - \frac{1}{2} \sum_{i,j}^n \alpha_i \alpha_j y_i y_j \mathbf{x}_i \cdot \mathbf{x}_j$$
[18]

subject to the constraints

$$\forall i \ \alpha_i \geq 0$$
$$\sum_i^n \alpha_i y_i = 0$$
[19]

where $\alpha_i$ are Lagrange multipliers to be determined. In a trained SVM, training vectors that correspond to $\alpha_i > 0$ are known as support vectors; these lie closest to the hyperplane and define the interclass boundaries.

For the non-separable case, a user-specified error penalty term $C$ is introduced, such that the constraints become

$$\forall i \ 0 \leq \alpha_i \leq C$$
$$\sum_i^n \alpha_i y_i = 0$$
[20]

In both cases, the solution weight vector is

$$\mathbf{w} = \sum_i^n \alpha_i y_i \mathbf{x}_i$$
[21]

The bias term $b$ can be chosen such that the hyperplane lies in the center of the margin.

We can now make comparisons between the linear SVM and the possibilistic linear classifier. The linear SVM normalizes $f(\mathbf{x})$ as in [16], while the possibilistic linear classifier normalizes the weight vector to unit length. The linear SVM maximizes the margin indirectly by minimizing the weight vector magnitude, as justified by structural risk minimization (Vapnik 1995). On the other hand, the possibilistic linear classifier maximizes the margin directly, as justified by uncertainty minimization.

Since the hyperplane that maximizes the interclass margin is unique, so obviously both the possibilistic linear classifier and the linear SVM find the *same solution* despite the different formulations. From a maximum-margin point of view, the linear SVM can be viewed as a possibilistic linear classifier.

Note however that in the non-separable case, the possibilistic linear classifier does not require an error penalty term. This is because the possibilistic solution is the hyperplane that minimizes the interclass overlap.

### 4.1 NONLINEAR CLASSIFIERS

The question arises as to whether possibilistic *nonlinear* classifiers would give better performance. For example, if we were to optimize the possibilistic criterion for a multi-layered perceptron network, would we get superior performance? In general, the answer is 'not necessarily'. Such a possibilistic nonlinear classifier might overfit the interclass boundary in order to maximize the margin width, thus resulting in poor generalization. Overfitting is also a phenomenon in probabilistic classifiers as well as in many other nonlinear statistical learning models. Some form of regularization or cross-validation is typically employed to reduce overfitting.

A nonlinear SVM circumvents this problem to a certain extent by first mapping each input vector to a feature space of very high (possibly infinite) dimension via a fixed mapping $\phi$, and then applying a linear SVM in feature space. The equation of the hyperplane (in feature space) is

$$f(\mathbf{x}) = \mathbf{w} \cdot \phi(\mathbf{x}) + b$$
[22]

In practice, we do not deal with the mapping $\phi$ explicitly, but rather through a kernel function $K$ such that

$$K(\mathbf{x}_i, \mathbf{x}_j) = \phi(\mathbf{x}_i) \cdot \phi(\mathbf{x}_j) \quad [23]$$

Examples of kernel functions include

Gaussian: $\quad K(\mathbf{x}_i, \mathbf{x}_j) = \exp\left(\dfrac{-|\mathbf{x}_i - \mathbf{x}_j|^2}{2\sigma^2}\right) \quad [24]$

Polynomial: $\quad K(\mathbf{x}_i, \mathbf{x}_j) = (\mathbf{x}_i \cdot \mathbf{x}_j + 1)^d \quad [25]$

where the kernel parameter $\sigma$ or $d$ is chosen by the user. The kernel parameter controls the extent to which the system fits the data. It should be chosen to avoid either overfitting or over-generalization.

The SVM solution for the hyperplane becomes

$$f(\mathbf{x}) = \sum_i^n \alpha_i y_i K(\mathbf{x}, \mathbf{x}_i) + b \quad [26]$$

where the coefficients $\alpha_i$ and the bias term $b$ can be found using quadratic programming as in the linear case.

We can adopt a similar approach in the possibilistic framework, by simply plugging the hyperplane equation [26] into the possibilistic likelihood function [14]. The possibilistic classifier would then maximize the interclass margin *in feature space*. Again, the same solution as that by the nonlinear SVM would be obtained.

### 4.2 ALTERNATIVE INTERPRETATIONS

We have not found any attempts in the literature that cast the SVM in a possibilistic setting. However, there has been some recent work on the probabilistic interpretation of SVMs (Platt 1999; Sollich 1999). In (Platt 1999), for example, the output of a trained SVM is fitted with a parametric sigmoid, which is then treated as a class conditional probability. In (Sollich 1999), on the other hand, SVMs are interpreted as maximum a posteriori solutions to inference problems with Gaussian process priors. As shown earlier, our possibilistic interpretation leads naturally to the SVM solution. Whatever the case, different interpretations are based on different premises and have different implications.

## 5 DISCUSSION AND CONCLUSION

We have proposed an uncertainty framework that is characterized by a generalized likelihood function. We have shown that given different uncertainty measures, maximizing this likelihood function induces different types of classifiers. It is interesting that the same framework under different conditions (probabilistic versus possibilistic) would give rise to two very different classification criteria. Among other things, the proposed uncertainty framework allows us to view different types of classifiers from a unified perspective.

If we can model the underlying probability distributions accurately and reliably, then it would be appropriate to use the probabilistic framework. Otherwise, the possibilistic framework might be a better alternative, since it makes weaker assumptions about the uncertainties involved.

There is a vast amount of literature on the probabilistic approach to classification (Devroye et al. 1996; Jurgen 1996; Langley et al 1992), as well as on the possibilistic approach (Bezdek et al 1999; Klir & Yuan 1995; Tanaka & Guo 1999). As far as we know, our treatment of uncertainty in the context of classification is distinct from those in the literature.

From another perspective, it is also interesting that two very different formulations would arrive at the same classification strategy. For example, the possibilistic classifier is based on minimizing the classification uncertainty, while the SVM is based on minimizing the structural risk. Their formulations are different. Yet, both give the same solution based on maximizing the interclass margin, either in input space for the linear case or in feature space for the nonlinear case.

In this paper, we have studied only a small part of the proposed uncertainty framework. Indeed, there is much more to explore, such as other conjunction operators, other uncertainty representations, the multi-class case, and the derivation of new learning algorithms.

### Acknowledgments

We would like to thank the reviewers for their valuable comments and suggestions.

### References

Baum E.B., & Wilczek F. (1988). Supervised learning of probability distributions by neural networks. In *Neural Information Processing Systems*, editor D.Z. Anderson, pp.52-61.

Bezdek J.C., Keller J., Krisnapuram R., & Pal N.R., (1999). *Fuzzy Models and Algorithms for Pattern Recognition and Image Processing*. Kluwer Academic Publishers.

Burges C.J.C. (1998). A tutorial on support vector machines for pattern recognition. *Data Mining and Knowledge Discovery*, 2, pp.121-167.

Cortes C., & Vapnik V. (1995). Support vector networks. *Machine Learning*, 20, pp.273-297.

Dempster A., Laird N., & Rubin D. (1977). Maximum likelihood from incomplete data via the EM algorithm. *Journal of the Royal Statistical Society*, B39, pp.1-38.

Devroye L., Gyorfi L., & Lugosi G. (1996). *A Probabilistic Theory of Pattern Recognition*. New York: Springer.

Gupta M.M., & Qi J. (1991). Theory of t-norms and fuzzy inference methods. *Fuzzy Sets and Systems, 40(3)*, pp.431-450.

Hertz, J., Krogh A., & Palmer R.G. (1991). *Introduction to the Theory of Neural Computation*. Addison-Wesley.

Hinton G.E. (1989). Connectionist learning procedures. *Artificial Intelligence*, 40, pp.185-234.

Jordan M., & Jacobs R. (1994). Hierarchical mixture of experts and the EM algorithm. *Neural Computation, 6*, pp.181-214.

Jurgen, S. (1996). *Pattern classification: a unified view of statistical and neural approaches*. New York: Wiley.

Klir G. J., & Yuan B. (1995). *Fuzzy Sets and Fuzzy Logic – Theory and Applications*, New Jersey: Prentice Hall.

Langley P., Iba W., & Thompson K. (1992). An analysis of Bayesian classifiers. *Proceedings of the Tenth National Conference on Artificial Intelligence (AAAI-92)*, pp. 223-228.

Platt J.C. (1999). Probabilistic outputs for support vector machines and comparisons to regularized likelihood methods. In *Advances in Large Margin Classifiers*, editors A. Smola, P. Bartlett, B. Scholkopf, & D. Schuurmans. Cambridge: MIT-Press.

Rosenblatt F. (1962). *Principles of Neurodynamics*. New York: Spartan.

Rumelhart D.E., Hinton G.E., & Williams R.J. (1986). "Learning internal representations by error propagation". In *Parallel Distributed Processing: Explorations in the Micro-structure of Cognition, Volume 1: Foundations*, editors D.E. Rumelhart, & J.L. McClelland, Chap.8, pp.318-362. MIT press, Cambridge.

Scholkopf B., Sung K., Burges C., Girosi F., Nigogi P., Poggio T., & Vapnik V. (1997). Comparing support vector machines with Gaussian kernels to radial basis function classifiers. *IEEE Transactions on Signal Processing, 45*, pp.2758-2765.

Sollich P. (1999). Probabilistic interpretations and Bayesian methods for support vector machines. *Proceedings of the Ninth International Conference on Artificial Neural Networks (ICANN'99)*, pp.91-96.

Tanaka H., & Guo P. (1999). Discriminant analysis based on possibility distributions. *Possibilistic Data Analysis for Operations Research*, Chap.7, pp.149-164. Heidelberg: Physica-Verlag.

Yang Y., & Liu X. (1999). A re-examination of text categorization methods. *Proceedings of the 22nd International Conference on R&D in Information Retrieval (SIGIR'99)*, pp.42-49.

Vapnik V. (1995). *The Nature of Statistical Learning Theory*. New York: Springer.